\documentclass{article}
\usepackage{amsmath,graphicx,mlspconf}
\usepackage{xcolor}
\usepackage{algorithm}
\usepackage{booktabs} % For formal tables
\usepackage{multirow} % To merge rows
\usepackage{float}  
\usepackage[noend]{algpseudocode}
\usepackage[skip=5pt]{caption} % Adjust the value (e.g., 5pt or 0pt) as needed
\usepackage{hyperref}
\usepackage{physics}
\usepackage{caption} 
\toappear{2025 IEEE International Workshop on Machine Learning for Signal Processing, Aug.\ 31-- Sep.\ 3, 2025, Istanbul, Turkey}

\title{Embedded Inter-Subject Variability in Adversarial Learning for Inertial Sensor-Based Human Activity Recognition}

 \name{%
    \begin{minipage}{\linewidth}
    \centering
     Francisco M. Calatrava-Nicolás$^{1,\star}$ %
     \qquad Shoko Miyauchi$^{1,\dagger}$ %
     \qquad Vitor Fortes Rey$^{3,4}$ %
     \qquad Paul Lukowicz$^{3,4}$  % 
     \qquad Todor Stoyanov$^{1}$%
     \qquad Oscar Martinez Mozos$^{5}$%
     \end{minipage}
     \thanks{$^{\star}$First Author - francisco.calatrava-nicolas@oru.se}%
     \thanks{$^{\dagger}$Corresponding Author - miyauchi@irvs.ait.kyushu-u.ac.jp}%
     \thanks{Francisco M. Calatrava-Nicolas and Todor Stoyanov gratefully acknowledge financial support from the Wallenberg AI, Autonomous Systems and Software Program (WASP) under the Knut and Alice Wallenberg Foundation. Computations were carried out on resources provided by the National Academic Infrastructure for Supercomputing in Sweden (NAISS), partially funded by the Swedish Research Council (grant no. 2022-06725).}% 
     \thanks{Vitor Fortes Rey and Pau Lukowicz also thank the Carl Zeiss Stiftung, Germany, for support through its Sustainable Embedded AI project (P2021-02-009) and the Cross-Act project (01IW25001).}%
      \thanks{ \textbf{This is the accepted manuscript. The final published article is available at \url{https://doi.org/10.1109/MLSP62443.2025.11204225}}}%
      \thanks{ © 2025 IEEE.  Personal use of this material is permitted.  Permission from IEEE must be obtained for all other uses, in any current or future media, including reprinting/republishing this material for advertising or promotional purposes, creating new collective works, for resale or redistribution to servers or lists, or reuse of any copyrighted component of this work in other works.}%
 }

  \address{%
      $^{1}$ Centre for Applied Autonomous Sensor Systems (AASS), Örebro University, Örebro, Sweden \\%
      $^{2}$ Faculty of Information Science \& Electrical Engineering, Kyushu University, \\ Fukuoka, Japan \\%
     $^{3}$ RPTU, Kaiserslautern, Germany \\%
     $^{4}$ DFKI, Kaiserslautern, Germany \\%
     $^{5}$ Escuela T\'ecnica Superior de Ingenier\'ia y Diseño Industrial, Universidad Polit\'ecnica de Madrid, \\ Madrid, Spain%
 }

\begin{document}
\ninept

\maketitle

\begin{abstract}
This paper addresses the problem of Human Activity Recognition (HAR) using data 
%collected 
from wearable inertial sensors.
%Human activity recognition (HAR) aims to classify human actions using data collected from sensors. 
%This approach finds applications in diverse areas, including 
%recognizing activities of daily living, monitoring well-being, and facilitating human-robot interaction. 
%shelathcare or sports.
% An important challenge in HAR is the inter-subject variability, i.e., the same activity is performed differently by different individuals.
An important challenge in HAR is the model's generalization capabilities to new unseen individuals due to inter-subject variability, i.e., the same activity is performed differently by different individuals.
%inter-subject variability, i.e, the same activity is performed differently by different individuals due to variations in personal styles and physical conditions. 
%This work addresses this problem by integrating 
To address this problem, we propose a novel deep adversarial framework that integrates the concept of inter-subject variability in the adversarial task, thereby encouraging subject‑invariant feature representations and enhancing the classification performance in the HAR problem.
%To address this problem, we integrate the concept of inter-subject variability into the adversarial learning framework to enhance the classification performance in HAR when using inertial sensor data. 
%This novel adversarial task compares a pair of signals—both recorded from the same activity—to determine if they originate from the same subject or from different subjects.
%we propose a novel adversarial task that explicitly focuses on discriminating common activities performed by different subjects. 
Our approach outperforms previous methods in three well-established HAR datasets using a leave-one-subject-out (LOSO) cross-validation. Further results indicate that our proposed adversarial task effectively reduces inter-subject variability among different users in the feature space, and it outperforms adversarial tasks from previous works when integrated into our framework. %Our approach improves the classification results when compared to previous methods on three well-stablished HAR datasets using a leave-one-subject-out (LOSO) cross-validation. Furthermore, our results indicate that our proposed adversarial task effectively reduces inter-subject variability among different users. Finally, we show that our novel inter-subject variability-based  adversarial task outperforms previous adversarial tasks 
%within the same adversarial framework, 
Code:\url{https://github.com/FranciscoCalatrava/EmbeddedSubjectVariability.git}
\end{abstract}

%\newcommand{\cem}[1]{\textcolor{blue}{cem: #1}}
%
%\vspace{-6}

\section{Introduction}
\label{sec:intro}

The problem of human activity recognition (HAR) consists of identifying the activities carried out by a person by analyzing the data collected with different sensors \cite{chaquet2013survey}. HAR is applied in different research areas, such as healthcare and well-being monitoring \cite{debes2016monitoring}, human-robot interaction (HRI) \cite{piyathilaka2015human}, or autonomous vehicles \cite{ohn2016looking}. 

This paper focuses on the application of HAR to classify activities using data from wearable inertial sensors. Although this problem has been lately addressed using different deep learning approaches 
\cite{tavares2023air,calatrava2023ieeesensors,yang2015deep,ShuSungho2022Percom,chen2020metier,bai2020adversarial,qin2023generalizable}, 
it still presents some challenges. A key issue, and the main focus of this paper, is to reduce the generalization gap 
%that appears when the model is evaluated in 
when applying the method to 
%evaluating 
previously unseen users. This gap arises from the heterogeneity in data distributions due to inter-subject variability, i.e., different individuals perform the same activity differently. In particular, 
individuals perform the same activity with varying levels of intensity and speed, influenced by factors such as personal preferences and physical characteristics. 
The resulting 
%temporal 
sensor-signal variations are often complex and non-linear, posing a challenge for accurate activity recognition 
%perception 
%by artificial systems 
\cite{barshan2016investigating}. This problem could be addressed by collecting and annotating extensive volumes of activity data to capture the diversity of real-world situations, however, it would require extra effort and resources. 
%\cite{Plotz2023Percom}.
%A potential solution for closing this generalization gap among individuals 
%due to inter-subject variability 
%would involve collecting and annotating extensive volumes of data from different people carrying out daily life activities to capture the diversity of real-world situations,
%mirror real-world conditions, 
%but this would require extra effort and resources due to collection and labeling. 
Therefore, it is desirable to design models that focus on learning features that generalize well among different individuals \cite{Plotz2023Percom}. 
%Recent research has aimed to reduce data distribution shifts among users to enhance classification performance without requiring additional data collection. In this context, several representation learning approaches—including multitask, adversarial, and self-supervised learning—have been proposed
Thus, recent research has focused
on reducing data distribution shifts among users to enhance classification performance without requiring additional data collection.
In this context, different representation learning approaches including
multitask, adversarial, and self-supervised learning have been proposed 
\cite{ShuSungho2022Percom,chen2020metier,bai2020adversarial,qin2023generalizable}.
These works still present some limitations. For example, \cite{chen2020metier} presents a multi-task learning framework that combines activity and user recognition, raising concerns related to privacy and scalability. 
Some methods attempt to address these 
privacy and generalization 
issues using adversarial learning, typically by introducing 
%various 
a user specific discrimination task within the adversarial framework \cite{bai2020adversarial, ShuSungho2022Percom}. However, these strategies often struggle to ensure consistent generalization over all users like in \cite{bai2020adversarial}, or they suffer from scalability constraints like in \cite{ShuSungho2022Percom}. A more detailed discussion of previous methods and their limitations is provided in Section~\ref{sec:related_work}.

 %These limitations 
 %challenges 
 %highlight the need for solutions that generalize well, scale better, and preserve privacy, a gap that this paper aims to narrow.

% However, these previous works have some limitations. For example, \cite{chen2020metier} 
% proposes a multitask learning framework based on activity and user recognition. Per contra , it faces scalability and privacy issues due to the integration of the user recognition task into the learning process. \cite{bai2020adversarial} mitigates the privacy issue by applying adversarial learning \cite{iwasawa2017privacy}. They propose a binary discrimination task based on classifying if a random pair of feature vectors comes from the same user. However, this random selection may undermine uniform generalization over all users. Similarly, \cite{ShuSungho2022Percom} proposes an user adversarial task with as many classes as users in the dataset, obtaining state-of the art results. However, it presents scalability issues as in \cite{chen2020metier}. Limitations from previous works are further discussed on Sec. \ref{sec:related_work}.

% To address these limitations while narrowing the generalization gap, we propose a new adversarial learning framework that integrates the concept of inter-subject variability.
%  To this end, we propose an adversarial learning framework that 
%  %propose to specifically 
% integrates the concept of inter-subject variability. 
To address these limitations 
%issues 
while narrowing the generalization gap, we propose a new adversarial learning framework that integrates the concept of inter-subject variability. We devise a novel adversarial task that determines whether a pair of activity feature vectors comes from the {\em same} person and the {\em same} activity, or from {\em different} persons but still the {\em same} activity. Our %primary
key
insight is that embedding the activity together with the user dimension within the discrimination task
%training discriminator dataset's construction logic 
will facilitate finding a common deep feature space for the same activity while intrinsically reducing the inter-subject variability for that activity, which helps close the generalization gap to unseen individuals. In addition, we frame our discrimination task as a binary classification problem, which does not increase the number of classes in the discriminator component with the number of users and thus enhances the scalability of our method. Moreover, we propose a novel combined loss function that integrates the non-saturating GAN loss function to learn subject-invariant feature representations. Finally, the adversarial nature of our framework may reduce privacy concerns \cite{iwasawa2017privacy}.
%In contrast to \cite{chen2020metier}, our adversarial task does not try to recognize specific users and, thus, reduces potential privacy issues. Moreover, our approach is a binary classification problem which does not increase the number of classes in the discriminator component and thus enhances the scalability of our method. Finally, and differently from \cite{bai2020adversarial}, our adversarial task adds the activity type, which allows a direct integration of the inter-subject variability of activities into the adversarial learning process. 
%To further integrate the inter-subject variability of activities into the adversarial learning process we reformulate the non-saturating GAN loss function \cite{goodfellow2016nips} in a way that effectively guides the feature extractor to produce embeddings in which features from different individuals performing the same activity appear indistinguishable from those of the same individual. %
%Furthermore, our model incorporates a new feature extractor based on the light residual network from \cite{calatrava2023ieeesensors}, which provided improved classification results in the HAR problem using a lighter architecture. 
%We evaluate our framework in a Leave-One-Subject-Out (LOSO) cross-validation setup ensuring that our model is tested against different new completely unseen individuals in  each iteration of the cross validation. 
%In addition, 
We evaluate our framework using a Leave-One-Subject-Out (LOSO) cross-validation setup in three well-established datasets (PAMAP2, MHEALTH, REALDISP) in the HAR field, ensuring that in each iteration of the cross-validation, the model is tested on data from a previously unseen individual.
LOSO is supported by previous works \cite{hammerla2015let,calatrava2023ieeesensors} as an effective method for simulating real-world conditions.
%Finally, we propose a novel combined loss function that integrates the non-saturating GAN loss function to learn subject-invariant feature representations.

In summary, our contributions are: (i) A novel deep adversarial framework that embeds the concept of inter-subject variability. (ii) Novel combined loss function to learn subject-invariant features. (iii) LOSO cross-validation comparison with state-of-the-art methods in three well-established datasets. (iv) Further ablation studies demonstrating the performance of our framework.

\begin{figure*}[t]
    \centering
%    \captionsetup{\textwidth}
        %\begin{adjustbox}{trim={0.02\width} {0.1\height} {0.02\width} {0.005\height},clip}
         \includegraphics[width=0.85\linewidth,keepaspectratio]{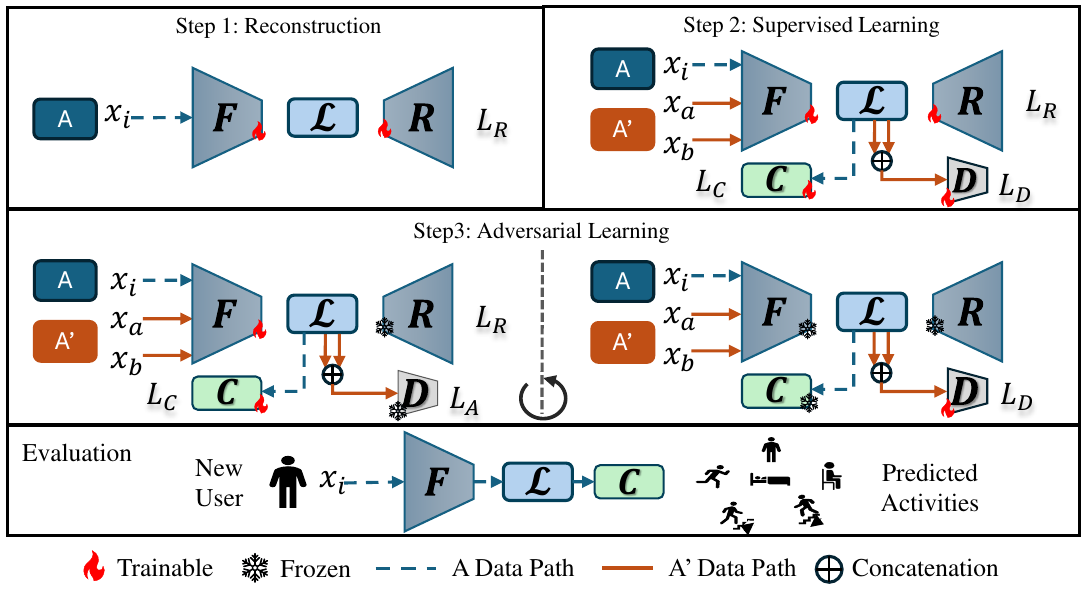}
        %\end{adjustbox}
        \caption{Overview of our framework: Steps 1–2 pre-train all module weights. Step 3 introduces adversarial learning to minimize inter-subject variability. Evaluation is performed on previously unseen users. Data distribution $A$ is used for the activity classification (Eq. \ref{eq:Adataset}), while $A'$ is the data distribution used in the adversarial task (Eq. \ref{eq:A_prime_definition}). 
        The circular arrow in Step 3 illustrates the iterative nature of adversarial learning.}
        % \caption{Our framework contains four modules: Feature Extractor $F$, Reconstructor $R$, Activity Classifier $C$, and Discriminator $D$. The framework is trained in three steps: Steps one and two are pretraining steps followed by the adversarial learning step. The blue curvy arrows in step three represents the iterative nature of the adversarial step. Blue and orange paths correspond to samples from $A$ and $A'$, respectively. Snowflakes denote frozen modules and flames denote trainable ones. The symbol $\oplus$ represents the concatenation operator.  At evaluation, the trained $F$ and $C$ alone perform activity recognition on unseen users.}
        %\hl{Please note that the testing phase assesses activity recognition performance with new, unseen individuals and thus only involves the feature extractor and classifier previously trained within the adversarial framework.}} 
        \label{fig:architecture}
\end{figure*}
%\vspace{-6mm}
%%%%%%%%%%%%%%%%%%%%%%%%%%%%%%%%%%%%%%%%%%%%%%%%
\section{Related Work}
\label{sec:related_work}
%%%%%%%%%%%%%%%%%%%%%%%%%%%%%%%%%%%%%%%%%%%%%%%%

%The HAR problem applied to the classification of activities using wearable inertial sensors has been lately addressed using different deep learning approaches.
The activity recognition problem using wearable inertial sensors has been
%lately 
addressed using different deep learning methods.
Convolutional neural networks (CNNs) have been used to classify windowed sensor signals as one-shot inputs \cite{calatrava2023ieeesensors,yang2015deep}, sometimes with attention mechanisms \cite{YangWang}. Recurrent models (RNN) like LSTM\cite{Fabio_Hernandez}
%and GRU \cite{Ullah} 
or hybrid models such as CNNLSTM \cite{ordonez2016deep} have been applied as well, showing good results. Multi-task learning has also been adopted to enhance performance across related tasks. For example, 
%\cite{peng2018aroma} distinguishes simple and complex activities, and 
\cite{chen2020metier} combines activity and user recognition in a multi-task setting. However, the complexity of the user classifier increases with the addition of new participants, raising concerns about scalability. Moreover, their use of cross-validation does not guarantee user-level separation for training and test sets, potentially leading to information leakage and an overestimation of the model’s ability to generalize to new users. In addition, \cite{chen2020metier} trains a user recognition network, which may lead to potential privacy issues \cite{iwasawa2017privacy}.
Recently, adversarial networks have been proposed to improve the HAR problem. These methods try to find a common feature space that generalizes among different users. This is achieved through an adversarial user-based discrimination task that guides the learning process towards less user-discriminant feature space that may reduce privacy concerns \cite{iwasawa2017privacy} and improve generalization. For example, \cite{bai2020adversarial} trains the discriminator to tell whether two feature vectors come from the same user, while simultaneously pushing the feature extractor to remove personal cues so that the discriminator can no longer tell them apart. However, to get these feature vectors, it samples user pairs randomly, which could hurt the uniform representation over all the users.
\cite{ShuSungho2022Percom}  improves this by assigning a unique class to each participant, achieving better generalization. However, the  user classifier also grows in complexity with the addition of new participants. 
%In this sense, the work in \cite{bai2020adversarial} proposes an adversarial binary discrimination task that determines whether two feature vectors come from the same participant or from different ones. The goal is to reach a feature space in the feature extractor that deceives the discriminator, making the features more invariant to individual users. However, this random selection method does not include all possible user combinations, which could limit its ability to generalize across all subjects.  To overcome that, the work in \cite{ShuSungho2022Percom} proposes an adversarial discrimination task in which a discriminator is trained to recognize each participant individually getting state-of-the-art results. In this case, however, the classifier also grows in complexity with the addition of new participants. 
Finally, the approach in \cite{qin2023generalizable} introduces self-supervised auxiliary tasks combining diversity generation and contrastive learning. While effective, their framework requires fine-tuning per test participant and uses a modified LOSO protocol, limiting comparability.
%still faces limitations regarding generalization, as it 
%requires 
%constant 
%fine-tuning of the trained model for each left-out set of testing participants.
%step of the 
%left-out
%independent testing dataset during evaluation.
%for different experiments 
%under the same dataset. 
%In addition, the authors apply a modified version of the LOOCV benchmark in which a reduced part of the available testing data is evaluated, and this limits direct comparisons with other methods. 
%the work , and its ability to generalize to new users remains uncertain due to its particular evaluation on a modified version of the Leave-One-Out cross-validation.
In contrast to the previous works, we propose a new adversarial framework that integrates a new discrimination task that takes into account the inter-person variability for the same activity. In particular, our discrimination task is framed as a binary classification problem that does not scale with the number of participants as in \cite{ShuSungho2022Percom}. Furthermore, and compared to \cite{bai2020adversarial}, we included the activity information in the discrimination task, proposing a tailored organization of the original dataset for this novel discrimination task. 
%and we re-structured the original dataset to improve the generalization among all the participants. 
Finally, our approach integrates a feature extractor with a reduced parameter count \cite{calatrava2023ieeesensors} to promote computational efficiency while preserving sufficient representational capacity.
\section{Adversarial Learning Framework for Inter-Subject Variability Reduction} 
\label{sec:method}

Our input set consist of signals from $c$ uni-dimensional sensor channels, segmented with a sliding window of size $w$. We write it as $X=\{x_i\}_{i=1}^M \ | \  x_i \in R^{w \times c} $,
% \begingroup
% \setlength{\abovedisplayskip}{6pt}
% \setlength{\belowdisplayskip}{6pt}
% \begin{align}
% X=\{x_i\}_{i=1}^M \ | \  x_i \in R^{w \times c} 
% \label{eq:Adataset}
% \end{align}
% \endgroup
where $M$ is the number of input data samples in the dataset. We then define our labeled input dataset for the activity recognition task as:
\begingroup
\setlength{\abovedisplayskip}{6pt}
\setlength{\belowdisplayskip}{6pt}
\begin{align}
A = \{x_i,y_i,s_i\}_{i=1}^{M} \  | \ y_i \in Y , s_i \in S
\label{eq:Adataset}
\end{align}
\endgroup
where $Y=\{y_1, \ldots, y_K\}$ represents the set of $K$ activity labels to be recognized, and $S$ is the set of subject. To integrate the concept of inter-subject variability,
%into our adversarial learning framework, 
we define a new input set $A'$ for 
%that will be used in 
our discrimination task. 
%This new set $A'$ is defined as:
%In this work, we additionally define a new input dataset $A'$ that will be used in our activity-based discrimination task as:
\begingroup
\setlength{\abovedisplayskip}{6pt}
\setlength{\belowdisplayskip}{6pt}
 \begin{align}
 A' {=} & \{ (x_a,x_b)^y_i, (s_a,s_b)_i, g_i \}_{i=1}^{N}\ | \nonumber \\  
    & x_a, x_b{\in}X,x_a{\ne}x_b, y{\in}Y, s_a,s_b{\in}S, g_i{\in}G{=}\{0,1\} 
 \label{eq:A_prime_definition}
 \end{align}
 \endgroup
 % \begin{align}
 % A' = & \{ (x_a,x_b)^y_i, g_i \}_{i=1}^{N} \ | \nonumber \\  
 %      & x_a, x_b \in X, x_a \ne x_b, y \in Y, g_i \in G=\{0,1\}
 % \end{align}

\noindent where $(x_a,x_b)^y$ represent two different data samples $x_a$,$x_b {\in} X$ that share the same activity label $y {\in} Y$; 
%In addition, $x_a$ is a feature vector that comes from subject $s_a \in S$, and $x_b$ is a feature vector that comes from subject $s_b \in S$.
$(s_a, s_b)$ are two subjects from which the corresponding data samples $(x_a$, $x_b)$ are selected. Finally, $g_i {\in} G$ is a binary label with value $1$ if data samples $x_a$ and $x_b$ belong to the same subject, i.e. $s_a = s_b$, or value $0$ if they belong to different subjects, i.e. $s_a \neq s_b$. 
%To address the fact that individuals may perform the same activity in slightly different ways, we construct a new dataset, $A'$, specifically tailored for our proposed adversarial discriminator task. 
The idea behind $A'$ is to account for the fact that different individuals may perform the same activity in 
%slightly 
different ways. 
%In $A'$, data samples are paired according to their activity labels, regardless of whether both data samples originate from the same subject or from different subjects (cf. Eq. \ref{eq:A_prime_definition}). 
%This pairing strategy is designed to enable the adversarial learning process to separate the intrinsic features of an activity from the variations caused by subject differences, such as variations in muscle recruitment patterns, movement speeds, or body postures that reflect each person’s unique physiology and habits \cite{barshan2016investigating,Plotz2023Percom}. 
Our main insight is that %incorporating 
embedding the activity label into the adversarial task encourages the model to learn a shared feature space for each activity, which in turn reduces the influence of inter-subject variability.
%The idea behind $A'$ is to incorporate the fact that different individuals may execute the same activity in slightly different ways. For instance, when performing a particular movement, each person might use different muscle recruitment patterns, movement speeds, or body postures based on their unique physiology and habits \cite{barshan2016investigating,Plotz2023Percom}. Therefore, we organize the new dataset $A'$ in pairs of common activities. Our primary insight is that embedding the activity within the training discriminator dataset will facilitate finding a common deep feature space for the same activity while intrinsically reducing the inter-subject variability for that activity.
%Because of these inter-person variations, we organize the dataset in pairs of 'same activity'. We think that this pairing approach might help to systematically capture how people perform the same activity differently, helping to the generalization towards new unseen individuals.
%The idea behind $A'$ is to take into account the inter-person variability, that is, different individuals implement the same activity in different ways. 
%The size of $A'$ grows with the potential combinations of pairs of feature vectors and pairs of subjects. Therefore, we reduce its size using a sampling strategy over the potential combinations while ensuring that each participant is adequately represented. In addition, we balance the binary distribution $G$ by keeping the same number of samples for each binary label.  
The size of $A'$ grows with the number of combinations of activities and subjects. To keep a manageable size, we uniformly sample the same number of data instances per class in $G$.

We define four blocks in our framework: the feature extractor $F$, the reconstructor $R$, the activity classifier $C$, and the discriminator $D$.  The architecture details for each module are provided in \cite{code}. The full framework is depicted in Fig. \ref{fig:architecture}. The feature extractor encodes the input sensor data $\mathcal{X}$ into a lower dimensional latent space $\mathcal{L}$, i.e. $F:\mathcal{X} \rightarrow \mathcal{L}$. 
%This means that ($batch_{size}$, 64) {\bf si  $x_i$ es una matriz, entonces ?como es el latent space?, ?una matriz tambien?}. 
The objective of $F$ is to learn a latent feature space that, while fooling the discriminator, remains effective for the activity classification task. In this work, we use the network from \cite{calatrava2023ieeesensors} without its classification layer.
%however, the framework is flexible and can accommodate other backbone architectures just as well.
% \begin{figure*}[t]
%     \centering
%     \captionsetup{width=\textwidth}
%         %\begin{adjustbox}{trim={0.02\width} {0.1\height} {0.02\width} {0.005\height},clip}
%          \includegraphics[width=0.5\textwidth]{Latex Template/new_figures/MLSPImage_v5.pdf}
%         %\end{adjustbox}
%         \caption{Framework}
%         % \caption{Our framework contains four modules: Feature Extractor $F$, Reconstructor $R$, Activity Classifier $C$, and Discriminator $D$. The framework is trained in three steps: Steps one and two are pretraining steps followed by the adversarial learning step. The blue curvy arrows in step three represents the iterative nature of the adversarial step. Blue and orange paths correspond to samples from $A$ and $A'$, respectively. Snowflakes denote frozen modules and flames denote trainable ones. The symbol $\oplus$ represents the concatenation operator.  At evaluation, the trained $F$ and $C$ alone perform activity recognition on unseen users.}
%         %\hl{Please note that the testing phase assesses activity recognition performance with new, unseen individuals and thus only involves the feature extractor and classifier previously trained within the adversarial framework.}} 
%         \label{fig:architecture}
% \end{figure*}
The reconstructor $R$ decodes the latent space $\mathcal{L}$ into the original input space $\mathcal{X}$, i.e. $R:\mathcal{L} \rightarrow \mathcal{X}$. 
%(architecture details in \cite{code})
This block stabilizes the adversarial training \cite{faridee2022strangan}.
%The parameters of the reconstructor are shown in Table \ref{tab:reconstructor_architecture}. 
We use the mean square error function for the reconstruction loss:
\begingroup
\setlength{\abovedisplayskip}{6pt}
\setlength{\belowdisplayskip}{6pt}
{
\small
\begin{align}
L_{R} = ||R(F(x_{i}))-x_i||_{2}^{2} 
\label{eq:LossReconstruction}
\end{align}
}
\endgroup
\noindent where $x_i$ is the input inertial signal (cf. Eq. \ref{eq:Adataset}). 

The activity classifier $C$ maps the latent space $\mathcal{L}$ into the activities space $\mathcal{Y}$, i.e., $C:\mathcal{L} \rightarrow \mathcal{Y}$. 
%(architecture details in \cite{code}).
%We employ a three-layer linear classifier of $64 \times 256$, $256 \times 512$, and $512 \times |Y|$ with a ReLU activation function between them. The last layer uses a softmax function to output a probability vector. 
%to minimize architectural complexity, ensuring that the performance reflects the quality of the learned latent feature representations in $F$ rather than the capacity of the classification network itself. It is composed of 3 fully connected layers of $64 \times 256$, $256 \times 512$, and $512 \times |Y|$ with a ReLU activation function between them. The last layer uses a softmax function to output a probability vector. 
%Since this is a multiclass classification problem, 
We use cross-entropy loss for the activity classification loss:
% \begin{align}
% L_{C} = -\sum_{c=1}^{C} y_c*log(C(F(x_i))_c)
% \end{align}
% \begin{align}
% L_{C} = -\sum_{\forall y \in Y} y*log(C(F(x_i)))
% \end{align}
 %\begin{align}
 %L_{C} = -\sum_{j=1}^{|Y|} z_j*log(C(F(x_i))_j)
 %\end{align}
 \begingroup
\setlength{\abovedisplayskip}{4pt}
\setlength{\belowdisplayskip}{4pt}
{
\small
 \begin{align}
 L_{C} = -\sum\nolimits_{j=1}^{|Y|} z_j \log C(F(x_i))_j
 \label{eq:LossClassification}
\end{align}
}
\endgroup

\noindent where $z$ is a binary vector of size $|Y|$  whose component $z_j=1$ if the original activity label for $x_i$ is $y_j$ or zero otherwise.

Lastly, our binary
discriminator $D$ maps the latent space ${\mathcal{L}}$ into our
activity-based binary discriminative classes $G$, i.e., $D:\mathcal{L} {\rightarrow} \mathcal{G}$.
%(architecture details in \cite{code})
$D$ takes as input the concatenated feature vectors of each pair drawn from $A'$
%(Eq. \ref{eq:A_prime_definition})
. We use binary cross-entropy as the discrimination loss:
 \begingroup
\setlength{\abovedisplayskip}{6pt}
\setlength{\belowdisplayskip}{6pt}
{
\small
\begin{align}
L_{D}= & -g_{i}\log D((F(x_{a})\oplus F(x_{b}))_{i})  \nonumber \\  
         & - (1{-}g_{i})\log 1{-}D((F(x_{a})\oplus F(x_{b}))_{i})
\label{eq:L_D}
\end{align}
}
\endgroup

\noindent where $x_a$ and $x_b$ represent the two data samples comprising the $i_{th}$ pair, $g_i$ corresponds to the label of the $i_{th}$ pair (cf. Eq. \ref{eq:A_prime_definition}), and $\oplus$ represents the  concatenation. 

Drawing inspiration from \cite{ShuSungho2022Percom}, our training process is divided into three steps. The first two steps focus on pre-training the weights of 
%the
%for the different 
%components 
the feature extractor $F$, the reconstructor $R$, the activity classifier $C$, and the discriminator $D$. The third step incorporates the adversarial learning.  
%aimed at guiding $F$ to produce a better user-invariant feature latent space that reduces the inter-person variability while is still meaningful for activity classification. 
Specifically, in step 1, we train $F$ and $R$ using the reconstruction loss $L_R$ on samples drawn from $A$ (Eq. \ref{eq:Adataset}).

%as:
%applying the following loss function:
% \begingroup
% \setlength{\abovedisplayskip}{6pt}
% \setlength{\belowdisplayskip}{6pt}
% \begin{align}
%     L_F^{step1} = L_R^{step1} = L_R
% \end{align}
% \endgroup
% In step 2, we train the framework in a multi-task way by training $F$, $R$, $C$, and $D$
% %all the models of our framework 
% at the same time. In this 
% %second 
% step, 
% %the feature extractor 
% %the reconstructor 
% $R$ and $D$ are trained by minimizing their respective losses $L_R$ and $L_D$ (Eq. \ref{eq:L_D}) on $A'$.
% %the reconstruction loss 
% The activity classifier $C$ is trained by minimizing 
% %the classification loss 
% $L_C$ (Eq. \ref{eq:LossClassification}) on $A$.
% % %the discriminator 
% % $D$ updates $\theta_D$ based o 
% % %the discriminator loss 
% % $L_D$ (Eq. \ref{eq:L_D}) on the distribution $A'$; 
% Finally, $F$ is trained by minimizing 
% %the reconstruction and classification losses 
% $L_R$ and $L_C$  using both $A'$ and $A$ data distributions. We defined the combined loss function for $F$ in step 2 as:
% %as follows:
% %applying the following loss function:
% \begingroup
% \setlength{\abovedisplayskip}{6pt}
% \setlength{\belowdisplayskip}{6pt}
% \begin{align}
%     L_{F}^{step2}(\theta_{F}{;}A,A'){=}L_C(\theta_F{;}A){+}L_R(\theta_F{;}A')
%     \label{eq:Lf_step2}
% \end{align}
% \endgroup

% \noindent where $\theta_F$ represents the parameters of $F$. 

In step 2, we train $F$, $R$, $C$, and $D$ simultaneously in a supervised way. In this step, $R$ and $D$ are optimized by minimizing $L_R$ and $L_D$, respectively, using data from $A'$ (Eq. \ref{eq:A_prime_definition}).  Additionally, $C$ is trained by minimizing $L_C$ on data from $A$. Finally, $F$ is updated by minimizing both $L_R$ and $L_C$, using data from $A$ and $A'$. We define the combined loss function for $F$ during step 2 as:
%as follows:
%applying the following loss function:
\begingroup
\setlength{\abovedisplayskip}{6pt}
\setlength{\belowdisplayskip}{6pt}
\begin{align}
    L_{F}^{\text{step2}}(\theta_{F}{;}A, A') = L_C(\theta_F{;}A) + L_R(\theta_F{;}A')
    \label{eq:Lf_step2}
\end{align}
\endgroup

\noindent where $\theta_F$ denotes the parameters of the feature extractor $F$.

% \begingroup
% \setlength{\abovedisplayskip}{6pt}
% \setlength{\belowdisplayskip}{6pt}
% \noindent
% \begin{minipage}{.5\linewidth}
% \begin{align}
%     L_F^{step2} &= L_R + L_C\\
%     L_R^{step2} &= L_R
% \end{align}
% \end{minipage}%
% \begin{minipage}{.5\linewidth}
% \begin{align}
%     L_C^{step2} &= L_C\\
%     L_D^{step2} &= L_D
% \end{align}
% \end{minipage}\\
% \endgroup

%Step 3 (deep adversarial learning) is divided into two sub-steps, inspired by the adversarial training paradigm in GANs \cite{goodfellow2016nips}:
The step 3 implements the  deep adversarial learning process. Inspired by the adversarial training paradigm in GANs \cite{goodfellow2016nips}, this third step is further subdivided into two sub-steps that are repeated iteratively. In the Step 3.1, $F$ is trained by minimizing a combined loss function consisting of $L_R$, $L_C$, and the adversarial loss $L_A$ (Eq. \ref{eq:Loss_A}) over data distributions $A$ and $A'$, while the weights of $D$ and $R$ are frozen. The combined loss function for $F$ in this step is:
\begingroup
\setlength{\abovedisplayskip}{6pt}
\setlength{\belowdisplayskip}{6pt}
    \begin{align}
    L_F^{\text{step3.1}}(\theta_F{;}A{,}A'){=}& w_AL_A(\theta_F{;}A') {+} w_RL_R(\theta_F{;}A')  \nonumber \\ 
    &{+} w_CL_C(\theta_F{;}A)
    \label{eq:loss3.1}
\end{align}
\endgroup
\noindent where $w_A$, $w_R$, and $w_C$ are scalar weights that balance the adversarial, reconstruction, and classification objectives, respectively, and $L_A$ is the non‐saturating loss \cite{goodfellow2016nips}. This novel combined loss unifies the adversarial, reconstruction, and classification terms, allowing $F$ to learn feature representations that preserve classification performance while fooling the discriminator. $L_A$ is defined as follow:
\begingroup
\setlength{\abovedisplayskip}{6pt}
\setlength{\belowdisplayskip}{6pt}
{
\small
\begin{align}
    L_{\scriptscriptstyle A}{=}{-}\log D((F(x_{\scriptscriptstyle a}){\oplus}F(x_{\scriptscriptstyle b}))_{\scriptscriptstyle i}),{\forall} \{(x_{\scriptscriptstyle a}{,}x_{\scriptscriptstyle b})_{\scriptscriptstyle i}{,}g_{\scriptscriptstyle i}\}{\in}A'{\mid} g_{\scriptscriptstyle i}{=}0
    \label{eq:Loss_A}
\end{align}
}
\endgroup

\noindent where $x_a$,$x_b$ represent the two data samples comprising the $i_{th}$ pair, $g_i$ corresponds to the label of the $i_{th}$ pair (cf. Eq. \ref{eq:A_prime_definition}), and $\oplus$ represents the concatenation operation.
%where $\oplus$ denotes the concatenation, and $g_i$ specifies the adversarial label (cf. \ref{eq:A_prime_definition}).
In Step 3.2, $D$ is trained by minimizing $L_D$ (Eq. \ref{eq:L_D}) using $A'$ as input set, while the parameters of $F$,$R$, and $C$ remain frozen.
% \begingroup
% \setlength{\abovedisplayskip}{6pt}
% \setlength{\belowdisplayskip}{6pt}
% \begin{align}
%     L_D^{step3.2} &{=} L_D
%     \label{eq:loss3.2}
% \end{align}
% \endgroup
This alternating optimization allows $F$ to learn latent feature representations that confuse $D$ (Step 3.1), while $D$ learns to distinguish them (Step 3.2), driving the adversarial learning. The intuition behind using the non-saturating adversarial loss (Eq. \ref{eq:Loss_A}) is to train the feature extractor $F$ to produce representations of data from different individuals, i.e., $g=0$ (Eq. \ref{eq:A_prime_definition}),  that, in the eyes of the discriminator $D$, seem to come from the same subject.
For that, the pair of feature vectors coming from a different user ($g=0$) are labeled as coming from the same subject ($g=1$). In this way, the feature extractor is pushed to remove subject-specific cues. The full training process is depicted in Algorithm \ref{alg:training}.

\begin{algorithm}[tb]
\small
\caption{Training algorithm for our framework}
\begin{algorithmic}[1] % The number tells where the line numbering should start
    \For{$epoch = 1$ to $Epochs_{step1}$} \Comment{Step 1}
        \For{each $batch$ in $X_A$ from $A$}
            \State $\theta_{F} \leftarrow \theta_{F} - \alpha_{F} \grad_{\theta_F} L_{R}(\theta_{F};X_A)$ \Comment{Eq. \ref{eq:LossReconstruction}}
            \State $\theta_{R} \leftarrow \theta_{R} -  \alpha_{R} \grad_{\theta_R} L_{R}(\theta_{R};X_A)$
        \EndFor
    \EndFor
    \For{$epoch = 1$ to $Epochs_{step2}$} \Comment{Step 2}
        \For{each batch in $A$ and $A'$}
            \State $\theta_{F} \leftarrow \theta_{F} -  \alpha_{F}\grad_{\theta_{F}} L_{F}^{\text{step2}}(\theta_{F};X_{A},X_{A'})$ \Comment{Eq. \ref{eq:Lf_step2}}
            \State $\theta_{R} \leftarrow \theta_{R} -  \alpha_{R}\grad_{\theta_{R}}L_{R}(\theta_{R};X_{A'})$
            \State $\theta_{C} \leftarrow \theta_{C} -  \alpha_{C}\grad_{\theta_{C}} L_{C}(\theta_{C};X_A)$ \Comment{Eq. \ref{eq:LossClassification}}
            \State $\theta_{D} \leftarrow \theta_{D} -  \alpha_{D}\grad_{\theta_{D}} L_{D}(\theta_{D};X_{A'})$ \Comment{Eq. \ref{eq:L_D}}
        \EndFor
    \EndFor
    \State We freeze R
    \For{$epoch = 1$ to $Epochs_{step3}$}\Comment{Step 3}
        \For{each batch in $A$ and $A'$}
            \State We freeze D (Step 3.1)
            \State $\theta_{F} \leftarrow \theta_{F} - \alpha_{F} \grad_{\theta_{F}} L_{F}^{\text{step3.1}}(\theta_{F};X_A,X_{A'})$ \Comment{Eq. \ref{eq:loss3.1}}
            \State $\theta_{C} \leftarrow \theta_{C} - \alpha_{C} \grad_{\theta_{C}} L_{C}(\theta_{C};X_A)$
            \State We unfreeze D and we freeze F and C (Step 3.2)
            \State $\theta_{D} \leftarrow \theta_{D} - \alpha_{D} \grad_{\theta_{D}} L_{D}(\theta_{D};X_{A'})$
        \EndFor
    \EndFor
\end{algorithmic}
\label{alg:training}
\end{algorithm}

\section{Datasets and Setup}
\label{sec:datasets}

We apply our method to three HAR datasets: PAMAP2 \cite{Reiss2012}, MHEALTH \cite{banos2014mhealthdroid}, and REALDISP  \cite{realdisp}. 
PAMAP2 ~\cite{Reiss2012} contains 54 sensor cues from 9 subjects performing 18 (12 from a protocol + 6 extra 
%activities
) activities. They provided inertial signals from 3 IMUs (worn on the dominant wrist, chest, and dominant leg). For classification, we use the 3D acceleration (±16g) and gyroscope data corresponding to the 12 protocol activities, excluding data from the ninth subject due to insufficient samples. We segment the data using a 512-sample sliding window with a 50\% overlap. MHEALTH ~\cite{banos2014mhealthdroid} comprises 12 activities from 10 participants with sensors on the chest, right wrist, and left ankle; we use the 3D acceleration and gyroscope segmented with a 512-sample window and a 50\% overlap. REALDISP ~\cite{realdisp} includes 33 activities from 17 participants using 9 sensors placed on various body parts; we use the accelerometer and gyroscope signals (54 signals total) and segment the data using a 256-sample sliding window with a 50\% overlap.Across all datasets, windows spanning multiple labels are discarded.

Following~\cite{hammerla2015let,calatrava2023ieeesensors}, we apply LOSO cross validation,
which consists of testing the model on one different user in each iteration, while training in the remaining set of users.
%ensuring no user overlap between train and test. 
In this way, test users remain unseen until evaluation. In addition, we designate two users from the training as the validation set to monitor the overfitting during training. All datasets are normalized via min-max scaling using parameters from the training sets. Each experiment is repeated twice with different random seeds, and the results are averaged.
%LOSO cross validation consists on running one experiment per user in the dataset, each time leaving out a different individual from the dataset to serve as the unseen test subject. This ensures that each user is evaluated with no data overlap between training and test sets. 
% %%%%%%%%%%%%%%%%%%%%%%%%%%%%%%%%%%%%%%%%%%%%%%%%%%%%%%
% \section{Setup}
% \label{sec:setup}
% %%%%%%%%%%%%%%%%%%%%%%%%%%%%%%%%%%%%%%%%%%%%%%%%%%%%%%
We compare our approach with prior works on the HAR field using wearable inertial sensors, in particular, 
%we compare with 
MCCNN \cite{yang2015deep}, DCLSTM \cite{ordonez2016deep}, METIER \cite{chen2020metier}, UIDFE  \cite{ShuSungho2022Percom}, and  DDLearn \cite{qin2023generalizable}.
%These methods are further explained in Sec. \ref{sec:related_work}. 
Note that while the original UIDFE method \cite{ShuSungho2022Percom}  incorporates unlabeled test data during training, we modify it to exclude any test data. 
\setlength{\textfloatsep}{12pt}  
\begin{table}[t]
\centering
\footnotesize
\caption{Comparison with previous approaches (Best in bold). 
%An asterisk (*) indicates that our proposed approach achieves a statistically significant improvement over the previous work ($p < 0.05$, t-test).
}
\begin{tabular}{p{1.3cm}p{1.2cm}cc}
\toprule
Dataset & Model &  Accuracy & $F1-Score_{M}$ \\
\midrule
 & MCCNN  &  $0.7939 \pm 0.0770$ &  $0.7511 \pm 0.0940$  \\
 & DCLSTM &  $0.7853 \pm 0.1012$ &  $0.7266 \pm 0.1113$  \\
 PAMAP2   & METIER  & $0.7924 \pm 0.0597$ & $0.7540 \pm 0.0574$  \\
 &  UIDFE &  $0.8014 \pm 0.1353$ &  $0.7648 \pm 0.1447$  \\
 &  DDLearn  &   $0.7188 \pm 0.1561$ &  $0.6442 \pm 0.1691$ \\
  &\textbf{Ours} & $\mathbf{0.8703 \pm 0.1219}$ &  $\mathbf{0.8643 \pm 0.1431}$ \\
\midrule
 & MCCNN  & $0.8378 \pm 0.0601$  & $0.6651 \pm 0.0998$\\
 & DCLSTM  & $0.9420 \pm 0.0497$  & $0.9094 \pm 0.0627$ \\
 REALDISP & METIER  & $0.9177 \pm 0.0567$  & $0.8099 \pm 0.1080$  \\
 &  UIDFE & $0.9450 \pm 0.0670$  & $0.9283 \pm 0.0700$  \\
 &  DDLearn  & $0.8476 \pm 0.0902$  & $0.7802 \pm 0.1323$  \\
& \textbf{Ours}  & $\mathbf{0.9710 \pm 0.0423}$  & $\mathbf{0.9651 \pm 0.0368}$ \\
 \midrule
 &MCsCNN  & $0.8736 \pm 0.0739$  & $0.7888 \pm 0.0823$  \\
 & DCLSTM & $0.8482 \pm 0.1051$  & $0.8157 \pm 0.1169$  \\
 MHEALTH & METIER  & $0.8583 \pm 0.0746$  & $0.8252 \pm 0.0932$ \\
 & UIDFE & $0.8982 \pm 0.0667$  & $0.8275 \pm 0.0881$  \\
 & DDLearn  & $0.8529 \pm 0.0684$  & $0.7825 \pm 0.0764$\\
 & \textbf{Ours}  & $\mathbf{0.9225 \pm 0.0606}$  & $\mathbf{0.9065 \pm 0.0780}$\\
\bottomrule
\end{tabular}
\label{tab:ComparisonSOTA}
\end{table}
Additionally, unlike DDLearn \cite{qin2023generalizable}—which uses only 20\% of left-out testing participants—we evaluate on 100\% of the test subjects. 
%This ensures consistency with the LOSO benchmarking and facilitates fair comparisons with previous works.
% It is important to note that the original UIDFE method incorporates unlabeled test data into the training process and thus, assumes that test data is available in advance. On the contrary, we assume no availability of any test data in advance and, therefore, we have modified the UIDFE method from \cite{ShuSungho2022Percom} to avoid the use of test data in the training process while the rest of the method remains the same. In addition, testing datasets in DDLearn \cite{qin2023generalizable} contains 
% %are structured into 
% %pairs or triples of individuals, and 
% only 20\% of the left-out testing participants. In our work, however, we use 100\% of the testing participants. This is done to be consistent with the LOSO benchmarking and to facilitate comparisons with previous works.
%, with one pair or triple reserved exclusively for testing. 
%However, only 20\% of this reserved test group is actually used for testing purposes. For instance, if there are four pairs, they would train four separate models, each using three pairs for training and the remaining pair for testing, but test only on 20\% of that remaining pair. To preserve as much as we can from this work, for DDLearn experiment we organise the datasets in pairs or triples, but we extend the testing to include the entirety of the data set left aside instead of only 20 \%.
%{\bf TODO: falta DDLearn.  FRAN: HECHO}
We implemented the methods from \cite{yang2015deep,ordonez2016deep,chen2020metier,ShuSungho2022Percom} according to their original descriptions, while for
\cite{qin2023generalizable}, we used the publicly provided code.
%\footnote{https://github.com/microsoft/robustlearn/tree/main/ddlearn}.
All experiments were run in Python 3.10.12 using PyTorch 2.2.0 on Ubuntu 18.04, running on NVIDIA Quadro RTX 6000 and NVIDIA Tesla A40 GPUs.
%The hyperparameters for our framework are presented in Table \ref{tab:hyperparameters}.
In Step 1, modules $F$ and $R$ are trained with a learning rate of $10^{- 4}$ for 20 epochs. Step 2 trains the same modules plus $C$ and $D$ for 10 epochs, with a learning rate of $10^{-4}$ for all except $C$, which uses $10^{-5}$. Finally, Step 3 further refines $F$, $C$, and $D$ for 150 epochs at a learning rate of $10^{-4}$ for all except $F$, which uses $10^{-3}$. We used Adam as optimizer.
% \begin{table}[t]
% \centering
% \caption{Hyperparameters for our framework}
% \begin{tabular}{llccc}
% \toprule
% & & Learning Rate    & Epochs & Optimizer \\
% \midrule
% Step 1 & $F$ & le-4  & 20     & Adam \\
%        & $R$ & le-4  & 20     & Adam \\
%        \midrule
% Step 2 & $F$ & le-4  & 10      & Adam \\
%  & $R$ & le-4  & 10      & Adam \\
%  & $C$ & le-4  & 10      & Adam \\
%  & $D$ & le-3  & 10      & Adam \\
%  \midrule
% Step 3 & $F$ & le-4  & 150    & Adam \\
%  & $C$ & le-4  & 150    & Adam \\
%  & $D$ & le-4  & 150    & Adam \\
% \bottomrule
% \end{tabular}
% \label{tab:hyperparameters}
% \end{table}
%. Those hyper-parameters together with the weights of the loss function $L_C^{step3.1}$ (Eq. \ref{eq:loss3.1}) were set empirically. 
The values for the weights $w_R{=}0.7$, $w_C{=}0.2$, and $w_A{=}0.1$ in Eq. \ref{eq:loss3.1} were found via grid search.
%also empirically set to 0.7, 0.2, and 0.1 respectively 
%using a grid search approach. 
All experiments were carried out under consistent hyperparameter settings and preprocessing
%, and segmentation procedures 
conditions (cf. Sec. \ref{sec:datasets}). 
%We additionally divided the datasets according to Sec. \ref{sec:datasets}.  
Moreover, the set $A'$ was resized to ensure an balanced number of samples for both classes in $G$: PAMAP2 and REALDISP contained 25,000 per class, and MHEALTH contained 5,000 per class (cf. Sec.\ref{sec:method}). The batch size for the $A$ set was 64 for step 1 and 32 for step 2 and 3. The batch sizes for $A'$ were adjusted to keep an equal number of batches with $A$.

%In addition, the set $A'$ has been resized (cf. Sec. \ref{sub:data_description}) for each dataset according to the total number of samples and the equality in terms of samples for both classes in $G$. For the PAMAP2, The $A'$ set contains 50,000 samples, with 25,000 samples for each binary class in $G=\{0, 1\}$. For MHEALTH, The $A'$ set consists of 10,000 samples, with 5,000 samples for each binary class in $G=\{0, 1\}$. Finally, for REALDISP, the $A'$ set contains a total of 50,000 samples, with 25,000 samples for each binary class in $G=\{0, 1\}$. The batch sizes for PAMAP2, MHEALTH, and REALDISP were set to 32 for $A$. The batch sizes for the $A'$ sets were adjusted to match the number of batches in the $A$ sets.

%The batch sizes for PAMAP2 were set to 64 for $A$ and 350 for $A'$. For MHEALTH the batch sizes were set to 32 for $A$ and 375 for $A'$. And for REALDISP they are 30 for $A$ and 395 for $A'$.

%\vspace{-6}
%%%%%%%%%%%%%%%%%%%%%%%%%%%%%%%%%%%%%%%%%%%%%%%%%%%%%%
\section{Experimental Results}
\label{sec:experiments}
%%%%%%%%%%%%%%%%%%%%%%%%%%%%%%%%%%%%%%%%%%%%%%%%%%%%%%

In this section, we first compare our proposed method
%, which integrates the concept of inter-subject variability, 
%with previous representation learning approaches that attempted the same, as well as with two basic baselines 
with previous works in the HAR field. We then conduct a distribution distance-based study, showing that our method reduces data shifts between train and test distributions. In addition, we provide an ablation study to measure the impact of each component in our framework. We also introduce a discrimination task study to compare discrimination tasks from previous works in our proposed framework. Finally, we provide a hyperparameter study for the weights in Eq. \ref{eq:loss3.1}. 
%Finally, we provide a hyperparameter study for the weights in the Eq. \ref{eq:loss3.1} to gain deeper insight into contributions of each part of the loss function.

%------------------------------------------------------
% \subsection{Comparison with Previous Approaches}

\begin{figure}[t]
    \centering
    \includegraphics[width=0.21\textwidth]{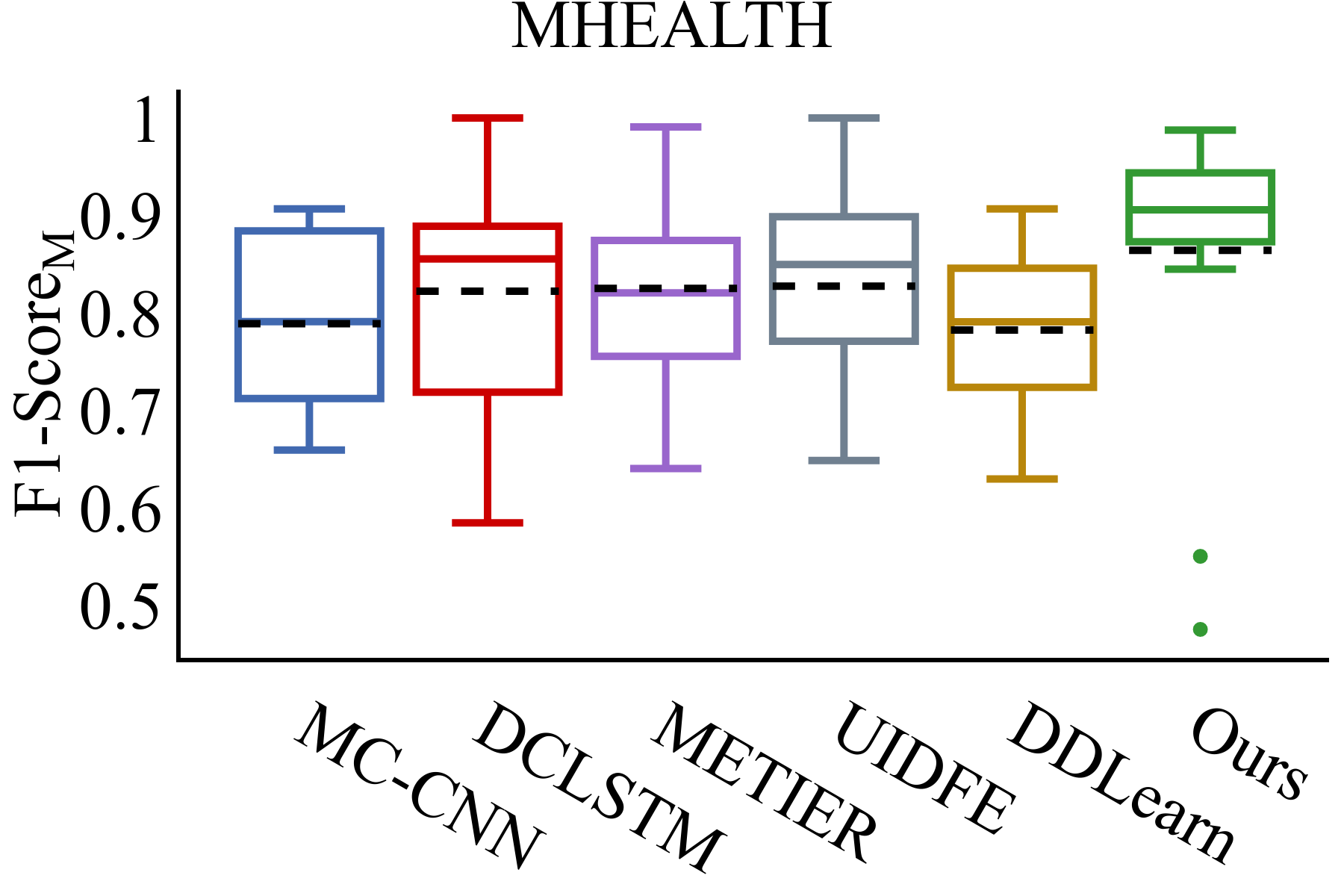}
    \hfill
    \includegraphics[width=0.21\textwidth]{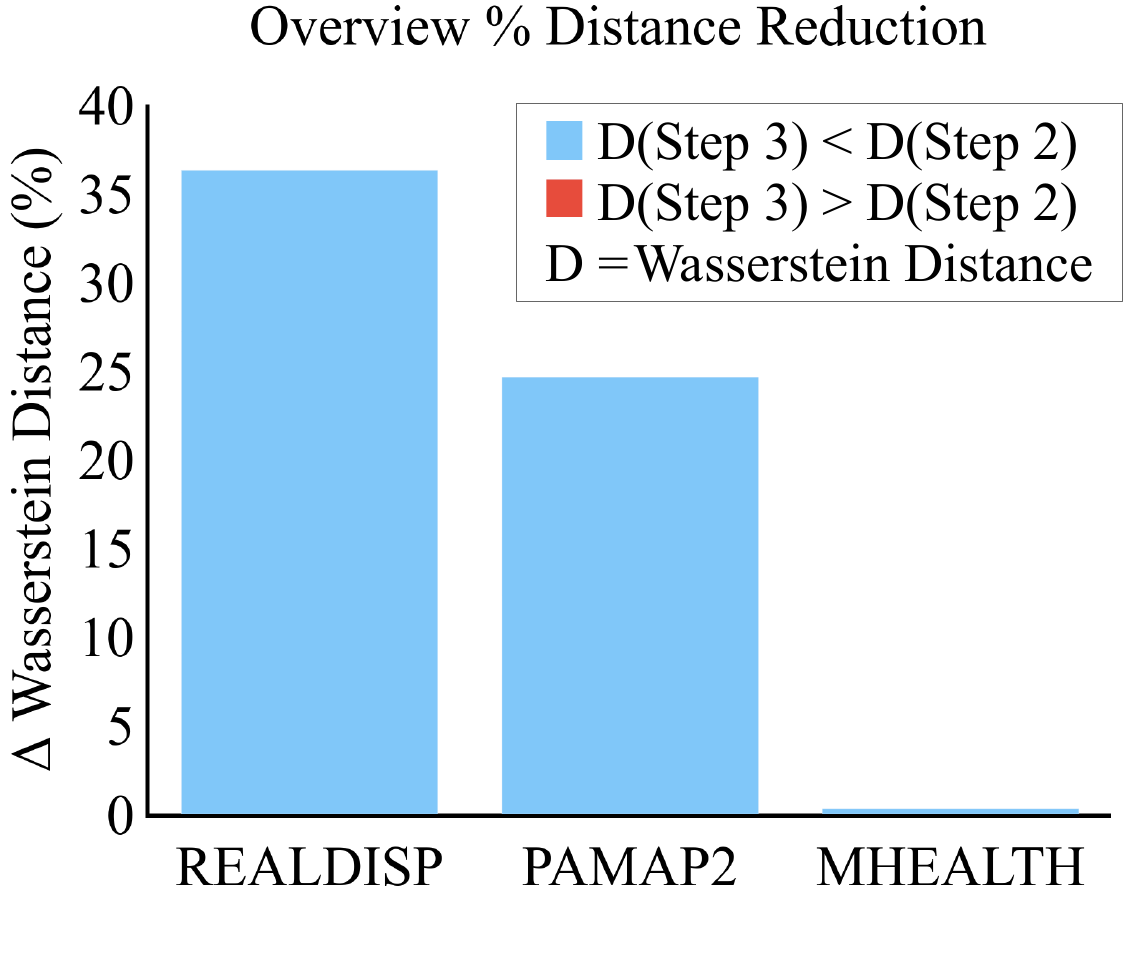}
    \\[2.5mm] 
    \includegraphics[width=0.21\textwidth]{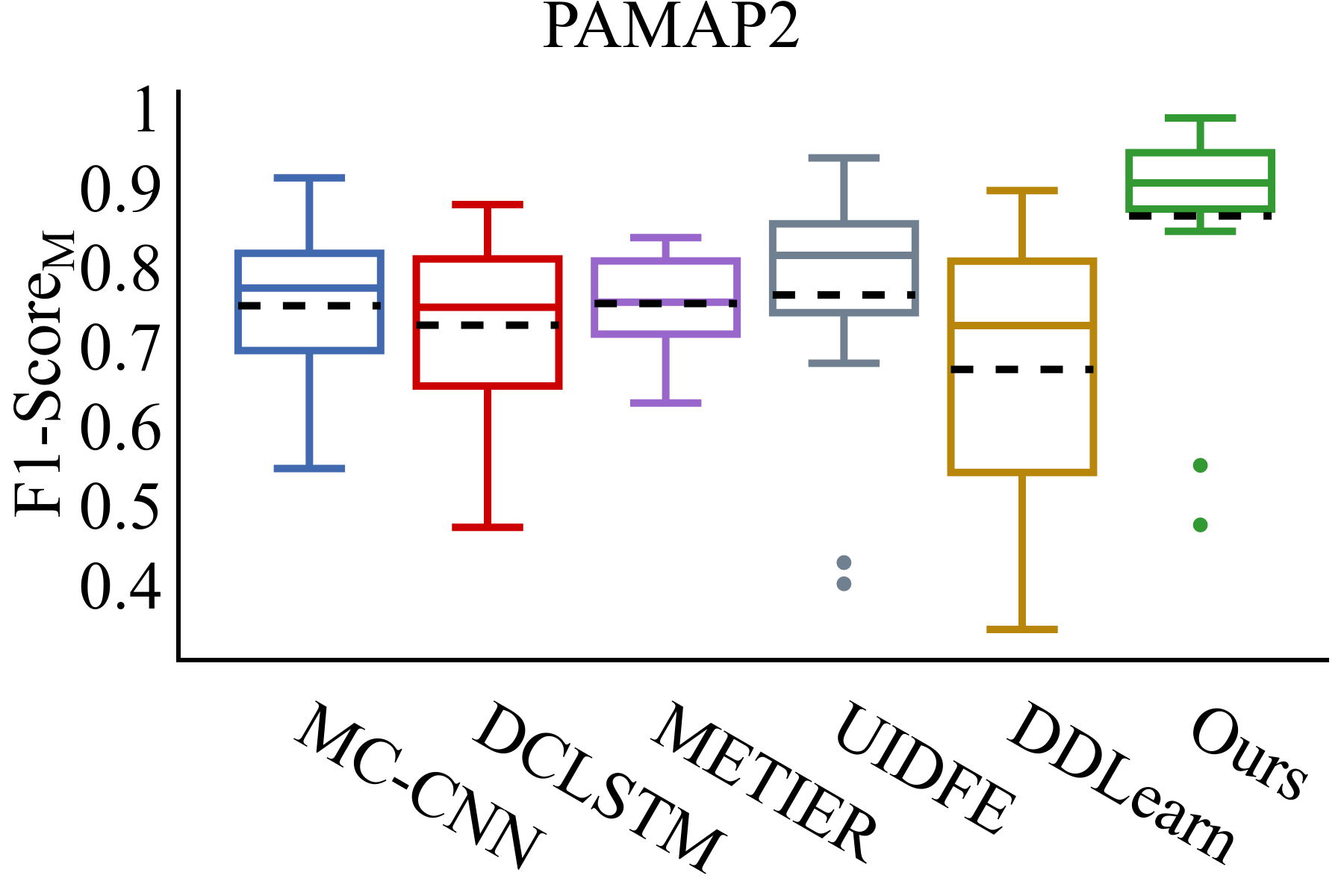}
    \hfill
    \includegraphics[width=0.21\textwidth]{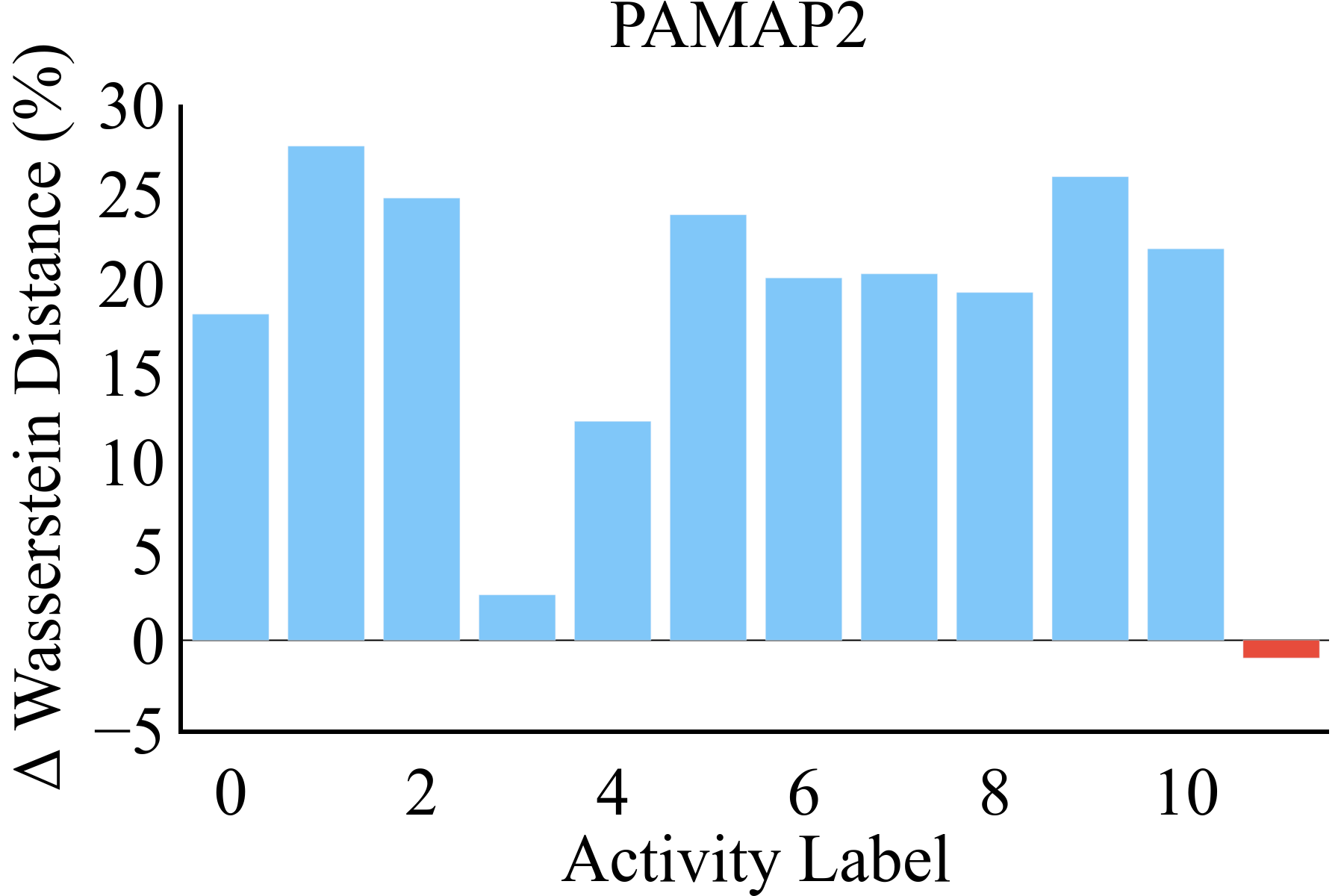}
    \\[2.5mm] 
    \includegraphics[width=0.21\textwidth]{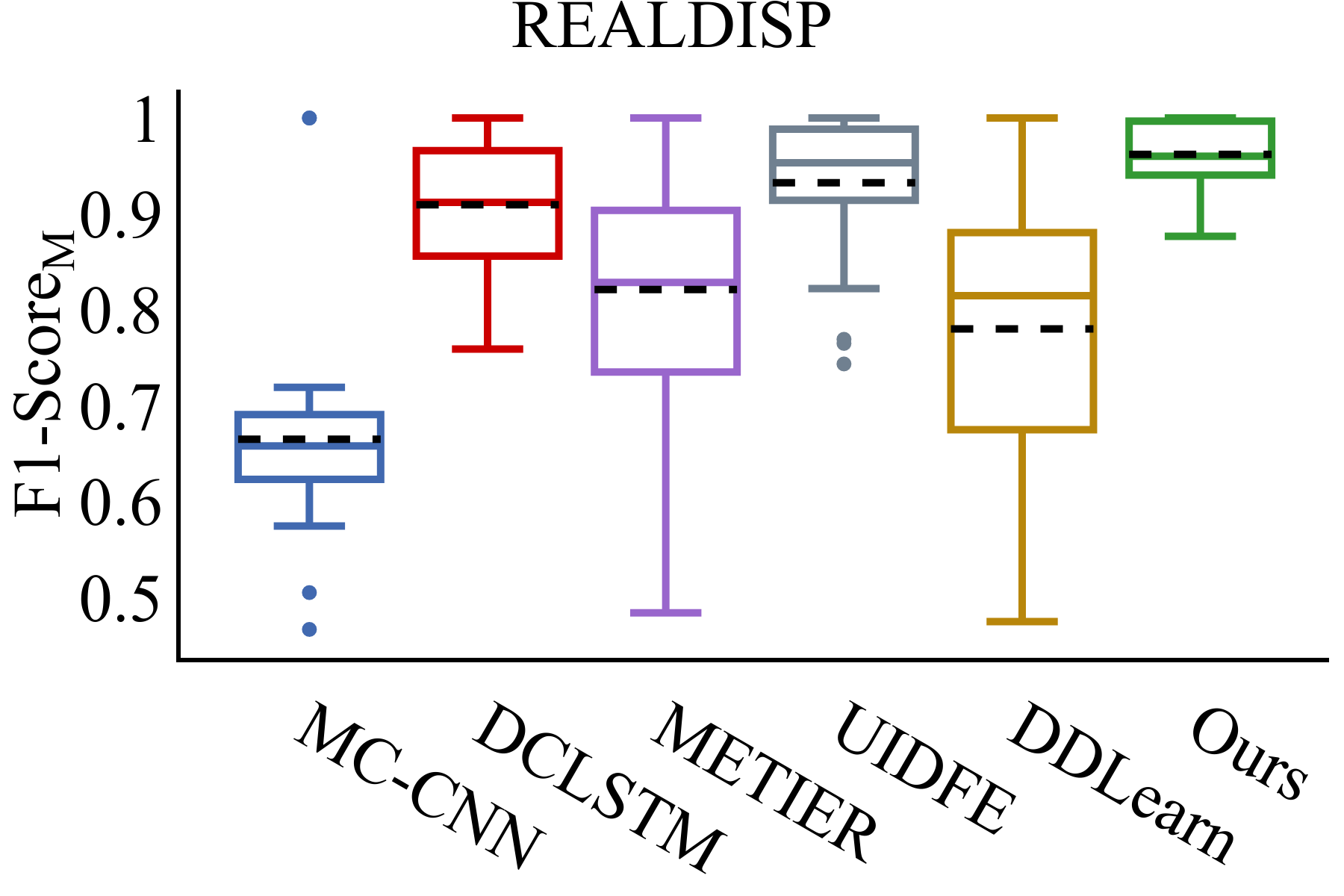}
    \hfill
    \includegraphics[width=0.21\textwidth]{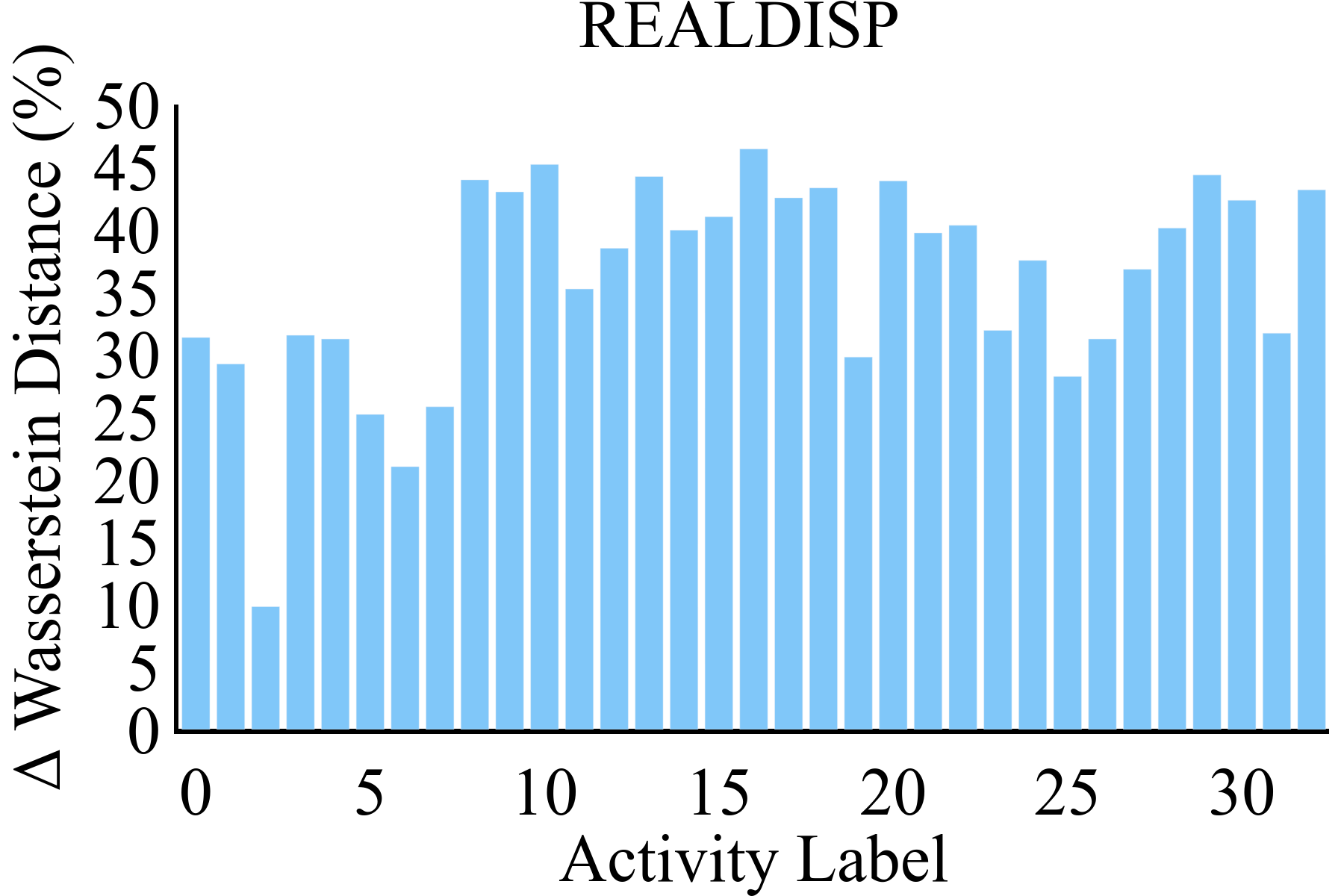}
    \caption{First column: $F1-Score_{M}$ for MHEALTH, PAMAP2, and REALDISP (top-down). Second column: Percentage change in Wasserstein distance from Step 2 to Step 3 between training and test set distributions. The top plot gives the distance change per dataset, while the middle (PAMAP2) and bottom (REALDISP) plots detail the distance change per activity. Positive bars (blue) mark a reduction (distance after Step 3 is smaller than after Step 2); negative bars (red) mark an increase.  
    %A positive percentage indicates decreased distribution discrepancy; a negative percentage indicates an increase. 
    %The center and lower bar plots depict the normalized Wasserstein distances of distributions for individual activities in PAMAP2 and REALDISP. 
    %The distances after Step 2 (grey diagonal pattern) and after Step 3 (blue no pattern) are normalized to sum up to 1. The black dashed horizontal line at 0.5 indicates equal distance values. Grey segments extending above 0.5 indicate a larger distance after Step 2 than after Step 3, showcasing a reduction due to our adversarial task.
    }
    %, which shows a reduction due to our adversarial task.} 
    \label{fig:combined_figures}
    %or below 0.5 indicate whether the Wasserstein distance is larger after Step 2 or Step 3, respectively. Same for blue ones. 
    %Wasserstein distance between train and test distributions for the same activity after Step 2 (grey) and after Step 3 (blue) in the training process (see the main text for more details). The left plot shows the average distance over all activities and all testing participants per dataset. The center and right stacked bar plots compare the normalized averaged Wasserstein distances for each activity for the PAMAP2 and REALDISP respectively. The distances after Step 2 and after Step 3 are normalized to sum up to 1. The black dashed horizontal line at 0.5 indicates equal distance values. Grey segments extending above 0.5 indicate a larger distance after Step 2 than after Step 3, which shows a reduction due to our adversarial task. %or below 0.5 indicate whether the Wasserstein distance is larger after Step 2 or Step 3, respectively. Same for blue ones.
    %In all cases, the grey bars (Step 2) are consistently larger than the blue bars (Step 3), reflecting greater distances after Step 2.
\end{figure}

First, we compare our method against previous works \cite{yang2015deep,ordonez2016deep,chen2020metier,ShuSungho2022Percom,qin2023generalizable} (Sec. \ref{sec:datasets}) using both Accuracy and F1-Score Macro ($F1-Score_{M}$). 
%The metrics were calculated with the Scikit-Learn package \cite{scikit-learn}.
As shown in Table \ref{tab:ComparisonSOTA}, our method outperforms all previous works under the same preprocessing conditions.
%Comparison results are shown in Table \ref{tab:ComparisonSOTA} where our  under the same conditions of preprocessing and segmentation.
%Moreover, our approach reduces most of the differences between the Accuracy and the two $F1-Score$ metrics. This trend suggests that our method remains more stable across diverse dataset compositions, and it demonstrates its ability to effectively handle classes with fewer samples without being overly influenced by the presence of dominant majority classes.
Fig. \ref{fig:combined_figures} (left column) further illustrates the distribution of F1-Score Macro using box plots.
%We also show results of the F1-Score Macro ($F1-Score_{M}$)   using box plots in Figure \ref{fig:combined_figures} (first column), 
Notably, our method yields the lowest Interquartile Range (IQR) in 2 of the 3 datasets, indicating reduced variability. In addition, our minimum score surpasses every other method’s lower bound across all datasets. Together, these findings suggest that our framework narrows the generalization gap and demonstrates stronger robustness, outperforming previous methods under a LOSO cross-validation setup.
Next, we explore the reduction of the inter-person variability when applying our approach. Specifically, we show that our adversarial task reduces the distribution distance for the same activity between training and testing participants. Our premise is that this reduction may facilitate the recognition of %the same 
activities in new unseen persons. To study this reduction, firstly, we measure the distribution distance for the same activity between the training and the left-out testing participant after Step 2 (Sec. \ref{sec:method}) and then we average over all left-out testing participants in the LOSO. Secondly, we repeat the same procedure after Step 3, which is the adversarial step. Finally, we compare both distances. 
We use the Wasserstein distance \cite{frogner2015learning} to measure the distribution distances. This comparison is done to study the effect of our adversarial task on the reduction of the inter-person variability for each activity.
Results are shown in Figure \ref{fig:combined_figures} (right column). In these plots, bars show the percentage change in Wasserstein distance from Step 2 to Step 3, expressed relative to the Step 2 distance. A positive bar means the distance decreased (distance after Step 3 is smaller that after Step 2), while a negative bar means it increased.
%In these plots, the Wasserstein distances after Step 2 (grey diagonal pattern) and after Step 3 (blue no pattern) are normalised to sum up to one. The black dashed horizontal line at 0.5 indicates equal distance values, and grey segments extending above 0.5 indicate a larger distance after Step 2 than after Step 3, which shows a reduction due to our adversarial task. 
The top plot in Figure \ref{fig:combined_figures} provides an overview of the decrease in Wasserstein distance from Step 2 to Step 3 per dataset. The results indicate a notable decrease in distance for the PAMAP2 and REALDISP datasets. This suggests that Step 3, which is the adversarial step, effectively reduces the discrepancy between training and testing distributions. For the MHEALTH dataset, the distance is marginally reduced, implying that the method has a limited impact in this case. These findings are in accordance with the results in Table \ref{tab:AblationStudy} (ablation study), which show greater improvements from Step 2 to Step 3 for PAMAP2 and REALDISP.
The center and lower plots of Figure \ref{fig:combined_figures} (right column) show specific results by activity for the PAMAP2 and REALDISP datasets, respectively. 
%We obtained the same trends with the Kullback–Leibler divergence metric.
%This means that if the distance between the train and the test is the same after steps 2 and 3, both bars (blue and grey) should meet on the dotted line. If the distance after step 3 is bigger than after step 2, the blue bar will surpass the dot line. In the opposite case, the grey bar will surpass the dotted line.  
%In the center and down plots from Figure \ref{fig:combined_figures}
Across both plots, distances after Step 3 are lower than after Step 2 for every activity, except for one PAMAP2 case where the distance increase is under 1 \%. This indicates that our adversarial discrimination task effectively reduces the inter-person variability for the same activities. 
%We do not provide detailed results for the MHEALTH dataset because the distances were very similar per activity, yielding the results shown in Figure \ref{fig:combined_figures} (right column, top). 
We omit detailed results for the MHEALTH dataset since the per‐activity distances after Step 2 and Step 3 were very similar, resulting in the overall improvement shown in Figure \ref{fig:combined_figures} (right column, top).

Following that, we present the ablation study of our framework. We compare three variants of our framework: one that trains $F$ together with a classification layer in an end-to-end supervised way as in \cite{calatrava2023ieeesensors}. Another that considers supervised learning together with reconstruction (essentially we stop training at step 2 in Algorithm~\ref{alg:training}). Finally, the third step presents the results of our full approach.
%Each of these experiments is run under identical conditions, including the data preprocessing, and is repeated with two different random seeds to check for consistency and to reduce the variance. 
The comparison results are shown in Table \ref{tab:AblationStudy}. By comparing results across these three experiments, we can assess how the reconstructor and the adversarial learning process each contribute to the final improved performance of our framework. Across all datasets, our proposed method consistently outperforms both baselines. These results suggest that the proposed adversarial step enhances the model's generalization capabilities.

\setlength{\textfloatsep}{12pt}  
\begin{table}[t]
\centering
\footnotesize
\caption{Ablation Study (Best in bold).}
\begin{tabular}{p{1.3cm}p{1cm}cc}
\toprule
 Dataset & Disc & Accuracy & $F1-Score_{M}$ \\
\midrule
 
        & Superv   &$0.7472 \pm 0.1067$  & $0.7444 \pm 0.1232$  \\
 PAMAP2 & Step 2& $0.8179 \pm 0.1255$ & $0.8229 \pm 0.1357$ \\
        &\textbf{Our}  & $\mathbf{0.8703 \pm 0.1219}$ & $\mathbf{0.8643 \pm 0.1431}$  \\
\midrule
         & Superv  & $0.9064 \pm 0.0724$ & $0.9220 \pm 0.0566$\\
REALDISP & Step 2 & $0.9372 \pm 0.0512$ & $0.9098 \pm 0.0620$ \\
         & \textbf{Our}& $\mathbf{0.9710 \pm 0.0423}$ & $\mathbf{0.9651 \pm 0.0368}$  \\
\midrule
         & Superv  & $0.9014 \pm 0.0633$ & $0.8934 \pm 0.0718$\\
  MHEALTH & Step 2 & $0.9121 \pm 0.0707$ & $0.8880 \pm 0.0891$\\
         & \textbf{Our} & $\mathbf{0.9225 \pm 0.0606}$ & $\mathbf{0.9065 \pm 0.0780}$  \\
\bottomrule
\end{tabular}
\label{tab:AblationStudy}
\end{table}

% %%%%%%%%%%%%%%%%%%%%%%%%%%%%%%%%%%%%%%%%%%%%%%%%%%%%%%
% \subsection{Comparison with Related Discrimination Tasks}
One of the main contributions of this paper is the discrimination task that integrates %implements
the concept of inter-person variability (Sec. \ref{sec:method}). To test its performance, we compare it to the discrimination tasks previously presented in \cite{ShuSungho2022Percom} and  \cite{bai2020adversarial}. To this end, 
%we use our proposed framework, in which 
we keep the same $F$, $R$, and $C$, and we change the discriminator $D$ 
accordingly 
%by implementing different 
to the discrimination tasks. We first implement a discriminator, $D_i$, that tries to identify each particular user as proposed in \cite{ShuSungho2022Percom}. 
%This discriminator outputs one class for each participant. 
The second implemented discriminator, $D_b$, takes as input two feature vectors and decides whether they belong to the same participant, as proposed in \cite{bai2020adversarial}. The third discriminator is our proposed task. Comparison results are shown in Table \ref{tab:discriminator_comparison}, where our discrimination task achieves the best classification results across all datasets. Moreover, it provides the smallest deviation for MHEALTH and REALDISP, and the second lowest for PAMAP2.

\setlength{\textfloatsep}{12pt}  

\begin{table}[t]   %figuras y tablas siempre [tbh], toop/bottom/here, por cuestion de estilo laas cosas arriba o abajo
\centering
\footnotesize
\caption{Comparison to prior discrimination tasks (Best in bold). }
\begin{tabular}{p{1.3cm}p{1cm}cc}
\toprule
 Dataset & Disc & Accuracy & $F1-Score_{M}$ \\
\midrule
\multirow{3}{*}{PAMAP2}
 & $D_{i}$ & $0.8274 \pm 0.1045$ & $0.7869 \pm 0.1102$\\
 & $D_{b}$ & $0.8130 \pm 0.1453$ & $0.7636 \pm 0.1579$ \\
 & \textbf{Ours} & $\mathbf{0.8703 \pm 0.1219}$ & $\mathbf{0.8643 \pm 0.1431}$ \\
\midrule
\multirow{3}{*}{REALDISP}
 &$D_{i}$ &$0.9447 \pm 0.0772$ & $0.9394 \pm 0.0566$\\
 &$D_{b}$ &$0.9511 \pm 0.0619$ & $0.9445 \pm 0.0538$ \\
&\textbf{Ours} & $\mathbf{0.9710 \pm 0.0423}$ & $\mathbf{0.9651 \pm 0.0368}$\\
\midrule
\multirow{3}{*}{MHEALTH}
 & $D_{i}$ & $0.9173 \pm 0.0698$ & $0.8892 \pm 0.0938$\\
 & $D_{b}$ & $0.9225 \pm 0.0707$ & $0.9010 \pm 0.0867$ \\
 & \textbf{Ours} & $\mathbf{0.9225 \pm 0.0622}$ & $\mathbf{0.9065 \pm 0.0801}$ \\
\bottomrule
\end{tabular}
\label{tab:discriminator_comparison}
\end{table}

\begin{figure}[t]
     \centering
    \includegraphics[width=0.9\columnwidth]{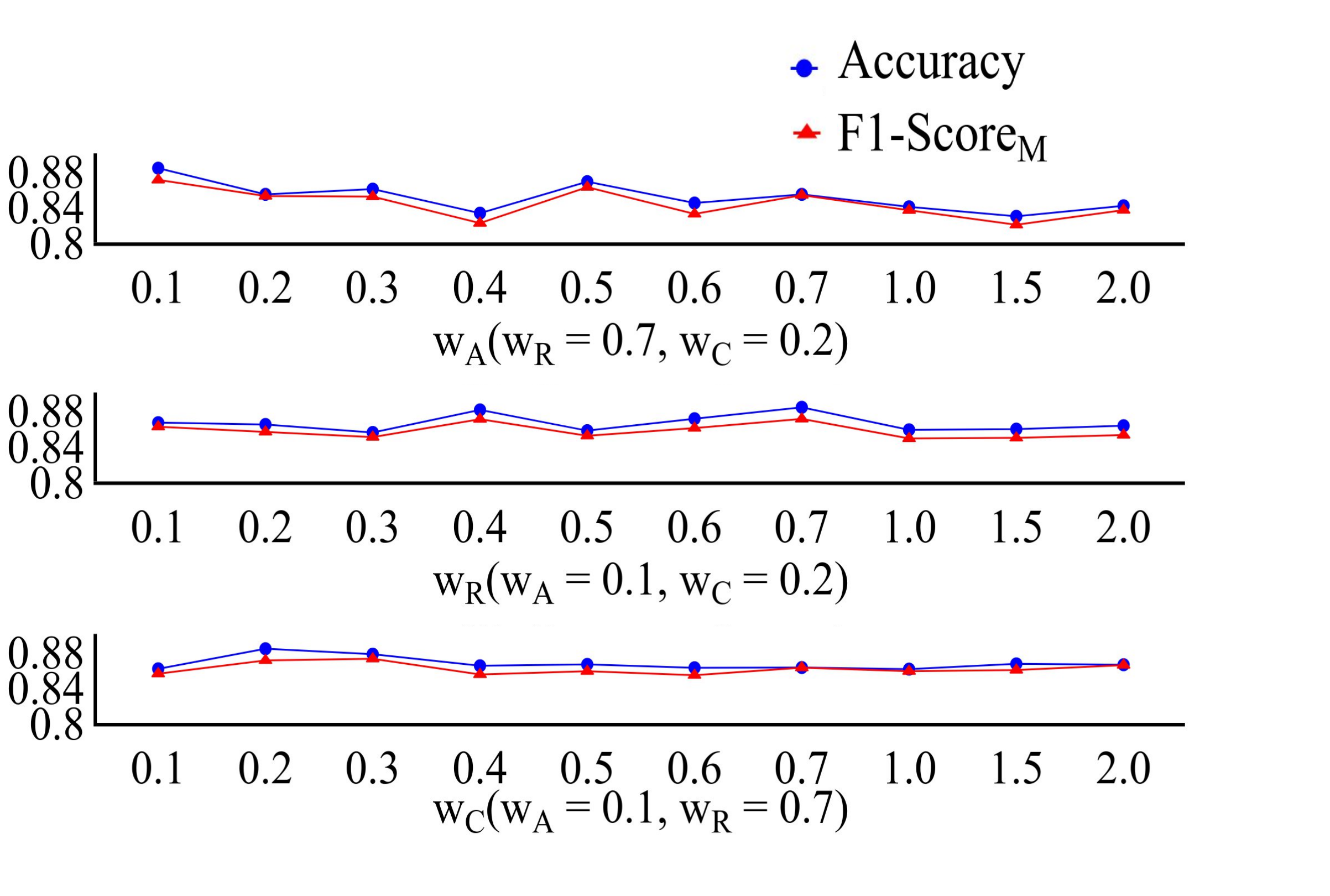}
    \setlength{\abovecaptionskip}{-10pt}
    \caption{Average accuracy and $F1-Score_M$ for variations in $w_A$ (top), $w_R$ (middle), and $w_C$ (bottom) for PAMAP2. 
    %Two weights remain fixed at values obtained from an initial grid search (Section \ref{sec:setup}), while the third weight is varied within a controlled range.
    }    
    \label{fig:weight_study}
\end{figure}

Finally, we investigate the influence of the three weights $w_R$, $w_C$, and $w_A$ in the combined loss function from Eq. \ref{eq:loss3.1} on the activity classification performance. The evaluation is based on the average classification accuracy and $F1-Score_M$ obtained from the LOSO benchmark when varying the range of a single weight while keeping the other two weights fixed. Two weights are held constant at the values determined through an initial grid search, as outlined in Section \ref{sec:datasets}, while the remaining weight is incrementally varied across a predefined range. Each data point presented in Figure \ref{fig:weight_study} corresponds to the mean classification accuracy obtained for a specific combination of the three weight values.
Figure \ref{fig:weight_study} shows that, when each weight is varied independently, the accuracy fluctuates by less than 1–2\% across the full range of values, indicating the model is largely insensitive to moderate shifts. 
%Moreover, the classification performance remains consistently high, showing that the trade-offs among the loss terms do not degrade the discriminative power. 
Overall, this highlights that our framework is robust to the choice of hyperparameter. 
%variations 
%in $w_R$, $w_c$, and $w_D$. 
%Consequently, for new tasks or datasets, one can adopt the default or grid-searched weights without significant risk of performance loss.

\section{Conclusions}
\label{sec:conclusion}
%%%%%%%%%%%%%%%%%%%%%%%%%%%%%%%%%%%%%%%%%%%%%%%%%%%%%%%%%%%%%%%%%%%%%%
We presented a new adversarial deep learning framework for the problem of human activity recognition based on wearable inertial sensor data. 
%on people wearing inertial sensors. 
Based on the concept of inter-subject variability, we designed a new discrimination task that takes into account information about the activity label together with the subject dimension, with the purpose of finding a common 
%deep 
latent feature space for the same activity among different subjects to reduce their distribution distances. 
%In addition, our adversarial model integrated the lighter feature extractor. 
%To validate the efficacy of our approach, 
We conducted extensive experiments on three distinct HAR datasets, using LOSO cross-validation to assess our proposed framework performance across unseen subjects.  The results show that our framework consistently outperforms existing state-of-the-art methods, highlighting its superior ability to generalize across different individuals. Furthermore, we performed comparative analyses against previously proposed discrimination tasks within adversarial learning paradigms in the HAR problem
%. These comparative analyses 
showing that our discrimination task improves the classification results when integrated into the adversarial framework.
%Our experimental results outperformed previous models in three activity datasets when using a LOSO cross-validation.
%An additional comparison with previously proposed discrimination tasks showed that our activity-based discrimination task improves the classification results when integrated into the adversarial framework.
%, and thus, we believe it is a good choice for adversarial models in the HAR problem. 
We also show evidence
%A further experiment shows evidence 
on the reduction of the inter-subject variability gap for the same activity between different users underscoring the potential of our approach to facilitate the development of more adaptable and user-agnostic HAR systems.
%Future work includes testing our model on larger datasets and implementing cross-dataset validation to better understand its robustness and generalization capabilities. 
%We will also continue to explore advanced data augmentation techniques such as synthetic data generation. These efforts will help improve our model's performance across diverse settings and further its applicability in human activity recognition.

\bibliographystyle{IEEEbib}
\bibliography{strings,refs}

\end{document}